# A Hybrid Data-Driven Algorithm for Real-Time Friction Force Estimation in Hydraulic Cylinders

1st Mohamad Amin Jamshidi
*Dept. of Mechanical Engineering*
*Amirkabir University of Technology*
Tehran, Iran
m-a.jamshidi@aut.ac.ir

2nd Mehrbod Zarifi
*Dept. of Mechanical Engineering*
*Amirkabir University of Technology*
Tehran, Iran
M.Zarifi79@aut.ac.ir

3rd Zolfa Anvari
*New Technologies Research Center*
*Amirkabir University of Technology*
Tehran, Iran
z.anvari@aut.ac.ir

4th Hamed Ghafarirad
*Dept. of Mechanical Engineering*
*Amirkabir University of Technology*
Tehran, Iran
Ghafarirad@aut.ac.ir

5th Mohammad Zareinejad
*New Technologies Research Center*
*Amirkabir University of Technology*
Tehran, Iran
mzare@aut.ac.ir

*Abstract*—Hydraulic systems are widely utilized in industrial applications due to their high force generation, precise control, and ability to function in harsh environments. Hydraulic cylinders, as actuators in these systems, apply force and position through the displacement of hydraulic fluid, but their operation is significantly influenced by friction force. Achieving precision in hydraulic cylinders requires an accurate friction model under various operating conditions. Existing analytical models, often derived from experimental data, necessitate the identification or estimation of influencing factors but are limited in adaptability and computational efficiency. This research introduces a data-driven, hybrid algorithm based on Long Short-Term Memory (LSTM) networks and Random Forests for nonlinear friction force estimation. The algorithm effectively combines feature detection and estimation processes using training data acquired from an experimental hydraulic test setup. It achieves a consistent and stable model error of less than 10% across diverse operating conditions and external load variations, ensuring robust performance in complex situations. The computational cost of the algorithm is 1.51 milliseconds per estimation, making it suitable for real-time applications. The proposed method addresses the limitations of analytical models by delivering high precision and computational efficiency. The algorithm's performance is validated through detailed analysis and experimental results, including direct comparisons with the LuGre model. The comparison highlights that while the LuGre model offers a theoretical foundation for friction modeling, its performance is limited by its inability to dynamically adjust to varying operational conditions of the hydraulic cylinder, further emphasizing the advantages of the proposed hybrid approach in real-time applications.

## I. Introduction

Hydraulic systems are known as an essential and extensively utilized system across multiple industries, operating through the application of incompressible fluids for power and energy transmission. The distinctive characteristics of these systems, such as their capacity to generate substantial forces while maintaining precision in motion control, make them effective in large and sensitive industries.
These capabilities are attributed to the incompressibility of fluids, meaning that even small changes in fluid volume can lead to significant pressure changes, which are subsequently transformed into mechanical motion.

Hydraulic actuators are widely used in industry, construction, transportation, and aerospace due to their significant power-to-weight ratio. Their capacity to generate great forces while providing precise control, coupled with the capability to function in adverse conditions, makes hydraulic actuators important. Hydraulic cylinders, a kind of hydraulic actuator in hydraulic systems, convert fluid power into motion. These essential components, used in many industries and machinery, produce great forces in a specified direction in linear motion. Precise modeling and analysis are crucial for hydraulic actuator devices to function reliably and safely, as well as improve performance and efficiency.

In hydraulic cylinders, the movement of the piston and fluid will be subject to friction. This friction is caused by the contact of the rod seals and piston seals with the cylinder and the viscous effects of the working fluid. Considering the properties of the seal, the amount of friction between the seal and the inner surface of the cylinder chamber will vary. Additionally, changes in the operating conditions of the hydraulic actuator, such as variations in the applied load, changes in fluid properties due to temperature changes and hydraulic fluid operation, changes in the environment, and actuator speed, will affect the friction force changes in the hydraulic cylinder. Changes in the friction of the hydraulic cylinder will affect the accuracy, controllability, and accuracy of the output applied by the hydraulic cylinder. Nonlinear friction in hydraulic systems, especially under low-speed conditions, can lead to reduced position tracking accuracy, increased tracking errors, and the emergence of phenomena such as oscillations and creep.

In recent years, some research has gone to developing models capable of precisely and consistently determine friction force in hydraulic and mechanical systems. The performance of mechanical systems is significantly influenced by friction force, which is a key factor in determining how well, how long, and how stable these systems are. It is essential to precisely identify and control friction because it can make control systems work much better.

### A. Analytical models

Nowadays, many analytical models exist to identify and predict friction behavior in mechanical systems. These models aim to explain characteristics like the Stribeck effect, presliding displacement, stick-slip motion, and frictional lag, which are commonly observed in real systems.

The Dahl model is one of the early models that represents friction as position-dependent, incorporating hysteresis effects. It assumes that friction is not just a function of the applied load but also depends on the history of the system's motion, providing a more accurate depiction of friction force in systems where hysteresis is significant, such as in dry contact and mechanical joints [1].

However, the Dahl model is limited in its ability to estimate dynamic friction accurately in systems where velocity-dependent effects are significant. This limitation led to the development of the LuGre model, a more advanced extension of the Dahl model. Developed by Canudas de Wit et al., the LuGre model introduces a dynamic state equation incorporating velocity dependence, offering a more refined representation of friction in sliding regimes. The model is particularly useful in mechanical systems, where friction must be controlled for precise motion, such as robotics and automotive systems. Despite its advantages, the basic form of the LuGre model has difficulty estimating the dynamic friction in systems like hydraulic cylinders when the system operates under oscillating velocity conditions. This makes it less effective for some applications, especially those involving high-speed or irregular motion [2]. An improved model of LuGre exists that incorporates the dynamics of lubricant films. Integrating lubricant film dynamics into the improved LuGre model provides several advantages over the traditional LuGre model. Specifically, it enhances the accuracy of simulating dynamic friction by effectively capturing the impact of fluid viscosity and type on friction behavior in hydraulic cylinders. This results in more reliable and precise estimations of friction force, leading to improved performance and efficiency in hydraulic systems operating under varying conditions [3].

*B.  Data-driven methods*

In recent years, the growth of machine learning and data-driven techniques, coupled with their efficacy in analyzing and modeling complex phenomena in mechanical engineering, has resulted in a significant rise in the use of these methods. Data-driven techniques in machine learning improve the detection of patterns and precise inference from many data points, allowing the accurate determination of friction force without the necessity for repetitive experiments and extensive calculations. Advanced algorithms such as artificial neural networks, support vector machines, and deep learning have been developed to estimate friction force. These algorithms, by utilizing extensive and complex data, are capable of learning nonlinear and intricate patterns and provide higher accuracy compared to classical methods. Data-driven methods and machine learning offer several advantages in estimating friction force. First, they do not require precise system modeling, as these methods can estimate friction force with high accuracy based solely on empirical data, without the need for complex mathematical models. Second, they are efficient and straightforward to implement, making them suitable for systems with diverse experimental data and allowing quick and reliable results. However, these methods also present some challenges. One issue is low sensitivity to noise, as experimental data may be affected by noise, which could reduce the accuracy of friction force estimation. Nevertheless, the data-driven approach mitigates this challenge by understanding the relationships between the model's inputs and outputs, which helps reduce sensitivity to noise. Another challenge is the need for extensive training data; achieving accurate and effective friction force estimates and training the algorithm requires a large volume of data related to system performance under various conditions.

Data-driven methods and machine learning have wide applications for estimating friction force in various uses of hydraulic cylinders. These methods, by increasing accuracy, enable the improvement of performance and the enhancement of safety in systems equipped with hydraulic cylinder actuators. Several studies that utilized a data-driven approach to determine the friction force in a hydraulic cylinder are presented below.

In 2006, Machado et al. [4] used an artificial neural network to compensate for friction in hydraulic cylinders. They designed a simple neural network architecture trained offline and capable of estimating the friction force online. This estimated force was then added as an input to the system's cascade controller to reduce the effects of friction in the actuator's motion path. This method used a variable friction coefficient model to train the neural network. The experimental results showed that this method was able to improve the performance of cylinder position control and reduce path tracking error compared to classical controllers. The Lyapunov stability method was also used to prove the stability of the closed-loop system.

In 2017, Pan et al. [5] introduced a new approach for modeling and identifying friction in hydraulic systems using an Elman Neural Network (ENN) optimized by a Multi-Agent Particle Swarm Optimization (MAPSO) combined with the Nelder-Mead Downhill Simplex (NMDS) algorithm. This method targets the nonlinear friction challenges in hydraulic systems at low speeds, which are critical for precision and stability. The approach not only refines the parameter estimation but also provides a more accurate estimation of friction forces compared to the traditional LuGre model. Experimental validations demonstrate the effectiveness of the ENN model, suggesting its potential for developing robust friction compensation controllers in hydraulic systems.

In 2023, Han et al. [6] developed a new model for predicting friction forces in hydraulic actuators using deep neural networks, trained on synthetic data from the LuGre model. This approach not only improved accuracy in friction prediction but also demonstrated efficiency in handling the complex, nonlinear characteristics of friction in industrial and aerospace hydraulic systems. The model's success hinges on the quality of training data, which, if not accurately representative, can impact the predictive performance. Despite the computational demands of training, the model offers significant potential for real-time applications once implemented. Further research is suggested to enhance the model's adaptability across different hydraulic systems and operational conditions.

This article attempts to provide a data-driven model for estimating friction force in hydraulic cylinders that can adapt to different operating conditions. These conditions include variations in pressure, velocity, temperature, lubrication state, and seal deformation, all of which significantly influence friction behavior. The model is going to be used online and real-time for estimating friction force based on signal after being trained with experimental setup.

## II. TEST SETUP AND EXPERIMENTS

### A. Test setup

The test setup used in this study is shown in Fig.1. A single-rod hydraulic cylinder (1) of model Bosch H160CA 63×28 was fixed horizontally on a frame (2). The motion of the hydraulic cylinder is controlled by a Festo 4/3-way hand lever valve. A Festo variable flow control valve was placed at the cylinder outlet to simulate an external load.

Two Trafag 0-250 bar pressure sensors (3) were used to measure the pressures, $P_1$ and $P_2$, in the cylinder chambers. The piston displacement, $x_p$, was measured using an Opkon linear potentiometer (4) coupled with the hydraulic cylinder.

Signals from the sensors were recorded into a computer through a 12-bit analog-to-digital (A/D) converter. Experimental data, such as displacement, $x_p$, and pressures, $P_1$ and $P_2$, were recorded at an interval of 5 milliseconds (200 Hz). To improve the quality of the measured data, a moving average filter with a window size of 10 was used to reduce the measurement noise.

The velocity, $\dot{x}_p$, of the piston was calculated by an approximate differentiation of the measured piston displacement. The noise in the calculated acceleration signal was filtered by a moving average filter with a window size of 30.

This setup is implemented in the Robotics and Servo Systems Laboratory at the New Technologies Research Center of Amirkabir University of Technology.

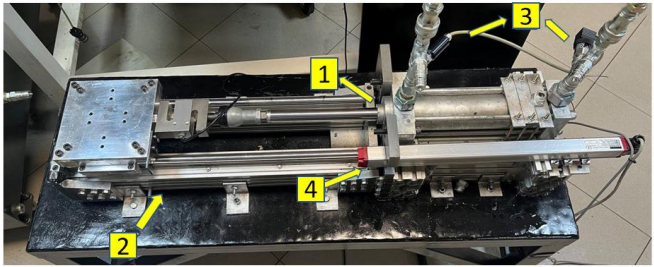

Fig. 1. Experiment platform, hydraulic cylinder, and test stand

### B. Schematic of test setup

Fig.2 shows the schematic of the test setup and the connection of the used elements including hydraulic cylinder (1), pressure sensors (3), linear potentiometer (4), power unit (5), relief valve (6), flow direction valve (7), one-way flow control valve (8) and data acquisition platform (9). The pressure sensors in the two chambers of the hydraulic cylinder, which record $P_1$ and $P_2$, are shown in the schematic. A linear potentiometer attached to the cylinder rod is also displayed for measuring position. The cylinder inlet is labeled as port $A$, and the outlet is designated as port $B$. Additionally, all flow paths connected to the variable displacement pump are marked with the symbol $P$, and all return lines connected to the tank are represented by the symbol $T$.

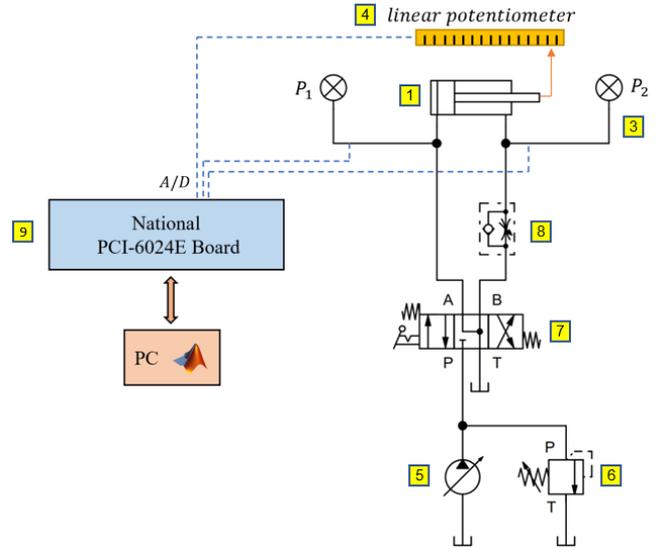

Fig. 2. Schematic diagram of the test setup

### C. Hydraulic actuator mathematical model

Fig.3 (a) shows two fluid chambers on the left and right, separated by a piston and a seal. The first chamber's inlet flow is $Q_1$, and the pressure on its side is $P_1$. The second chamber's inlet flow is $Q_2$, and the pressure is $P_2$. The piston contact area on the first chamber side is $A_1$, and the piston contact area on the second chamber side, eliminating the cylinder rod contact area, is $A_2$.

The friction force, $F_f$, is computed using the hydraulic piston's equation of motion, which takes the measured pressures, acceleration, and load mass effect. The effect of hydraulic fluid compressibility in the motion equation is represented by an equivalent spring stiffness. Fig.3 (b) shows the modeling of fluid compressibility as an equivalent spring in each chamber.

In fluid compressibility modeling, the equivalent spring stiffness $K_1$ and $K_2$ will be represented by equations (1) and (2).

$$K_1 = \frac{\beta A_1}{L_1} \quad (1)$$

$$K_2 = \frac{\beta A_2}{L_2} \quad (2)$$

where β is the bulk modulus of the fluid and $L_i$ is the distance of the piston from the cylinder wall.

Fig.3 (c) & (d) shows the free body diagram of the piston and rod and the forces acting on them during sliding to the right and left side. In Fig.3 (c) & (d), the force applied to each side of the piston is calculated by multiplying the chamber pressure by the contact area, and the nonlinear friction force, which is dependent on the piston's movement direction, is equal to $F_f$.

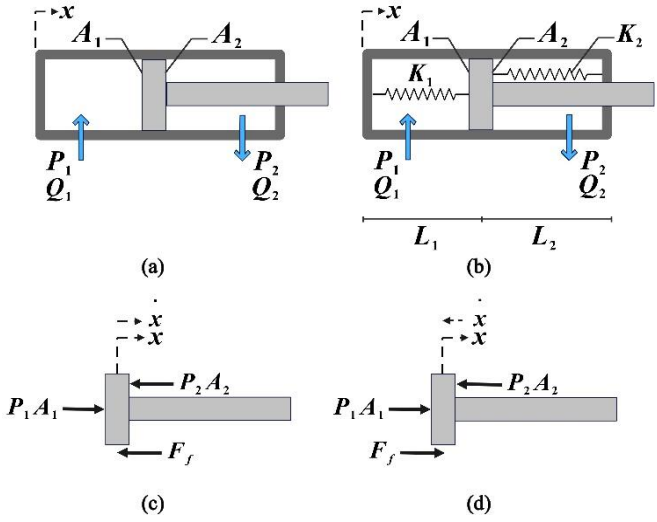

Fig. 3. (a) Schematic diagram of the hydraulic cylinder
(b) Schematic of hydraulic cylinder and equivalent stiffness
(c) Free body diagram of the piston and acting forces
in sliding to the right
(d) Free body diagram of the piston and acting forces
in sliding to the left

Equation of motion in sliding to the right:

$$m\ddot{x}_p = P_1A_1 - P_2A_2 - K_{eq}x_p - F_f \qquad (3)$$

Equation of motion in sliding to the left:

$$m\ddot{x}_p = P_2A_2 - P_1A_1 - K_{eq}x_p - F_f \qquad (4)$$

In equations (3) and (4), $\ddot{x}_p$ is the piston acceleration, $x_p$ is the piston displacement, and $K_{eq}$ is the equivalent stiffness derived from the fluid compressibility modeling.

## III. FRICTION FORCE ESTIMATION

### A. Hybrid Data-Driven Algorithm for Real-Time Friction Force Estimation

In this study, the friction force in the hydraulic cylinder is estimated online using a hybrid data-driven algorithm. As the data is time-series, understanding the dynamic relationships between the data is crucial for developing an adaptive model. The research aims to estimate the nonlinear friction force using a data-driven approach, requiring an algorithm to learn from the training dataset and accurately estimate the friction force for each dataset after training.

Furthermore, the research data is recorded at specific time intervals, showing that it is a time series. These data will be used to process a dynamic model, which necessitates considering the dynamic relationship between the model's results. The output of dynamic models is more than just a one-to-one mapping of the model's inputs; it also includes the model's previous predictions. As a result, the model must be able to identify relationships between data points and store the data. That is, the model's decision should be based on memory and the training dataset used during the training process.

Recurrent neural networks are a machine learning method that performs well at identifying relationships between training data. Because of this, recurrent networks are widely used in mechanical engineering applications. Hydraulic cylinders and similar mechanical systems exhibit behaviors influenced by their inherent mechanical memory. These systems often show time delays in response due to fluid dynamics and demonstrate nonlinear behaviors that result in complex and unpredictable responses. Environmental and operational changes further impact these dynamics, embedding long-term dependencies within the system's data. However, simple recurrent networks are not capable of learning long-term dependencies between data, and using simple recurrent networks increases prediction error [7].

To compensate for this limitation, a new algorithm based on recurrent neural networks called Long Short-Term Memory (LSTM) is proposed. These networks are particularly well-suited for dynamic data because they are designed to capture temporal dependencies and long-range correlations in sequences. Unlike traditional neural networks, LSTMs have specialized memory cells that can retain information over long periods, making them effective in handling time-series or sequential data with complex temporal patterns. This ability allows LSTMs to manage dynamic data where the current state depends not just on the immediate past but also on the historical context [8].

Considering the proposed algorithm's need to learn long-term dependencies between training data, a Long Short-Term Memory (LSTM) neural network is used in part of the proposed algorithm.

For the estimation algorithm, the random forest algorithm is recommended. A long short-term memory neural network extracts the relationships between the input data, which then goes into the random forest algorithm. The Random Forest algorithm is a nonlinear regression and classification algorithm in machine learning. It is commonly used in estimation and prediction models due to its processing speed, resistance to overfitting, and stability against noise and unorganized data. The Random Forest algorithm is based on decision trees and includes several decision-making units. The input data for each decision unit is randomly chosen from the input training set. Finally, the output of each decision unit is considered an estimate, and the final output of the random forest model is calculated by averaging the outputs of each unit [9].

To achieve the optimal number of blocks and layers of the long short-term memory network and the number of decision units in the random forest algorithm, using the hyperparameter optimization algorithm by the Optuna library in Python is suggested [10]. This algorithm seeks to minimize the cost function, which in the current study is the mean absolute error, and investigates various combinations of the model's hyperparameters, such as the number of layers in the long short-term memory network, the number of units in the long short-term memory network layer, the learning rate, and the number of units in the random forest processor, to determine the best combination for the network architecture. The mean absolute error cost function is calculated using equation (5).

$$MAE = \frac{1}{n}\sum_{i=0}^{n}|y_i - \hat{y}_i| \qquad (5)$$

where $n$ is the number of data points, $y_i$ is the calculated data value, and $\hat{y}_i$ is the estimated value by the algorithm.

Fig.4 shows the proposed algorithm and data flow in the model.

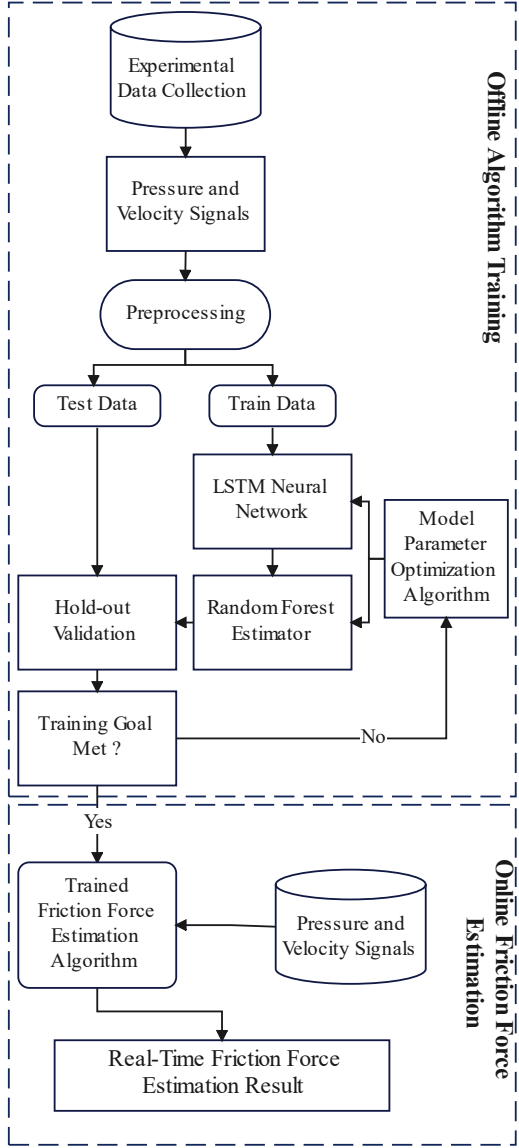

Fig. 4. Schematic of proposed algorithm and data flow

### B. Environment

The data acquisition environment was set up in Simulink using MATLAB. The algorithm was developed in Anaconda using Python 3. Several Python libraries were utilized for data processing, modeling, and evaluation. Pandas and Numpy were used for data manipulation and analysis. Visualization was performed using matplotlib and seaborn. Data preprocessing and model training were conducted using scikit-learn for splitting the data, standardizing features, and evaluating performance with the mean absolute error (MAE) metric. Machine learning models, including Random Forest and LSTM networks, were built using TensorFlow with Keras, with optimization performed using the Adam optimizer. Early stopping was implemented to prevent overfitting.

## IV. RESULTS AND DISCUSSION

### A. Training the algorithm

By running the algorithm and aiming for the lowest mean absolute error, the process of determining and training the network architecture begins. The optimization algorithm employed in my model is Adam. It iteratively adjusts the weights and biases of the algorithm to minimize the cost function. Adam employs adaptive learning rates to facilitate efficient convergence to the function's minimum [11].

The result of the hyperparameter optimization algorithm determines the network architecture. Table I shows the range of hyperparameter looks and the final value of each.

TABLE I. HYPERPARAMETER OPTIMIZATION VALUES

| Parameter | Value | |
|---|---|---|
| | *Range* | *Final Value* |
| LSTM Layers | $3, 4, 5$ | $4$ |
| LSTM Nodes per Layer | $16, 32, 64$ | $64, 64, 64$ |
| Dropout Rate | $0.1, 0.2, 0.3, 0.4, 0.5$ | $0.3$ |
| Learning Rate | $10^{-5} - 10^{-3}$ | $0.001$ |
| Random Forest Estimators | $32 - 100$ | $64$ |
| Random Forest Maximum Depth | $10 - 50$ | $10$ |

Fig.5 shows the training process of the algorithm. The mean absolute error decreases in each epoch until it reaches a minimum value. The performance of the proposed algorithm was evaluated using the hold-out validation method, where the dataset was partitioned into distinct training and testing subsets. This approach ensures that the evaluation is conducted on unseen data, providing a reliable estimate of the model's generalization capability, to validate the algorithm accurately, several independent sets of pressure and velocity signal data were applied to the model for validation, resulting in a gap between the training process error and the validation process.

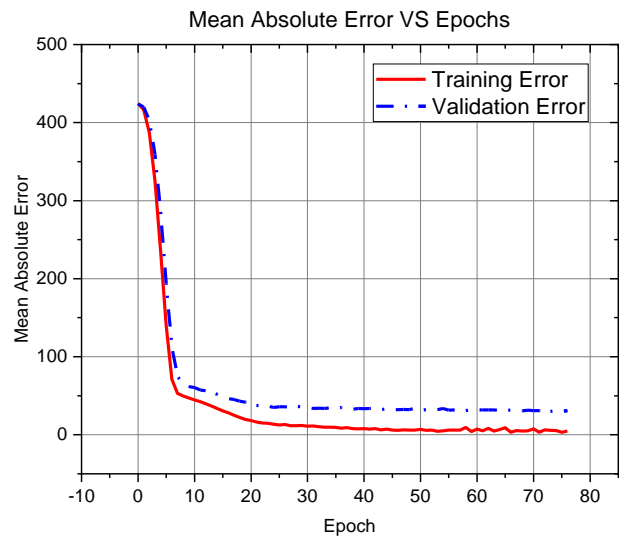

Fig. 5. Mean absolute error change in model training and validation

### B. Comparison with the LuGre model

To compare the efficiency of the proposed algorithm against existing analytical models, the algorithm's adaptability under diverse operational conditions of the hydraulic cylinder will be compared with the LuGre analytical model. A flow control valve at the hydraulic cylinder's output port is used to gradually regulate the output flow in four tests to simulate different loaded operating conditions. The external load's effect on the hydraulic cylinder will be apparent as an increase in pressure signals. The hydraulic cylinder's no-load conditions are simulated in Test 1, and the external load is progressively and incrementally simulated in Tests 2 to 4.

Fig.6 shows the hydraulic cylinder's friction force for the calculated, the proposed algorithm, and the LuGre analytical model.

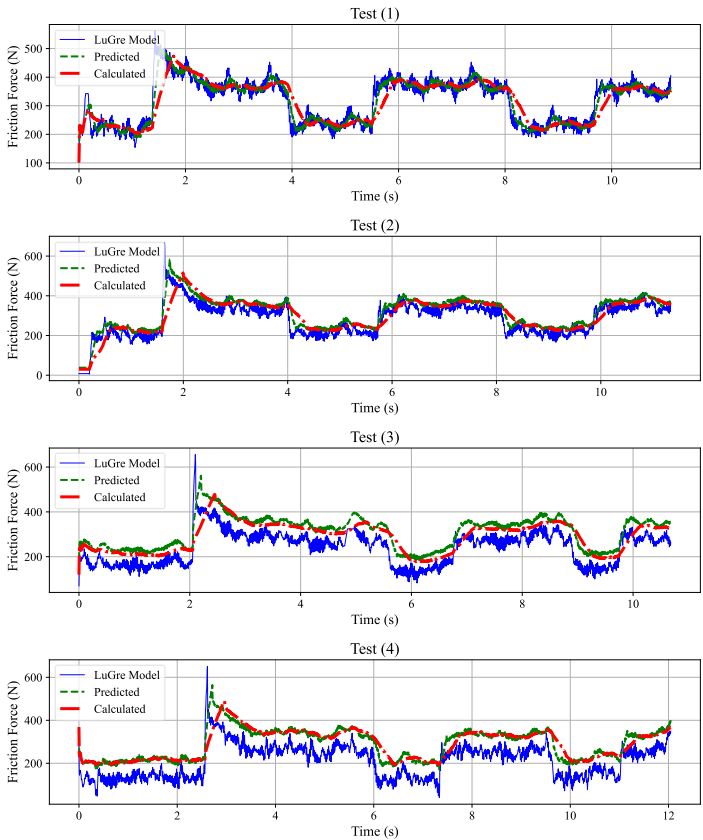

Fig. 6. The friction force with respect to time for the proposed model, the LuGre model, and the calculated approach

As the external load on the hydraulic cylinder increases, the error in the LuGre model analysis grows. The difference occurs because the parameters of the LuGre model are determined under no-load conditions and must be updated to reflect changes in operating conditions. But in reality, it becomes challenging to re-identify the model parameters as the external load on the hydraulic cylinder increases. The error values of the proposed algorithm and the LuGre analytical model evaluating changes to operating conditions are displayed in Table II.

TABLE II.    COMPARISON OF ERROR IN THE LUGRE MODEL AND THE PROPOSED ALGORITHM

| Test | Mean Absolute Error | |
|---|---|---|
| | *LuGre model* | *Algorithm* |
| Test (1)  [ No-Load] | 9.04 % | 5.76 % |
| Test (2) | 11.86 % | 6.35 % |
| Test (3) | 19.86 % | 8.84 % |
| Test (4) | 28.5 % | 5.57 % |

### C. Model performance

To examine if the model has been able to find the relationships between the training data and provide a new estimate of the friction force based on it, the distribution of the algorithm's residuals must be reviewed. The residual in the algorithm is calculated using equation (6).

$$e_i = y_i - \hat{y}_i \qquad (6)$$

where $y_i$ is the actual data value and $\hat{y}_i$ is the estimated value by the algorithm.

By plotting the residuals histogram during the algorithm's validation process, a proper analysis of the algorithm's ability to discover relationships between the data and the impact of each data point's error on the error of other data points can be performed. Fig.7 shows a histogram of the proposed algorithm's residual distribution.

The residuals are approximately normally distributed, indicating that the algorithm effectively captured the relationships between the data and had the ability to utilize the hidden relations in the data during the training process, despite the noise in the training data. Furthermore, the normal distribution of the residuals indicates that the errors in the model are independent. So the error of one estimation does not influence the error of another in the algorithm. The median of the residuals is close to zero, implying that the model does not overestimate or underestimate. The first and third quartiles show a symmetric distribution of residuals around the median value.

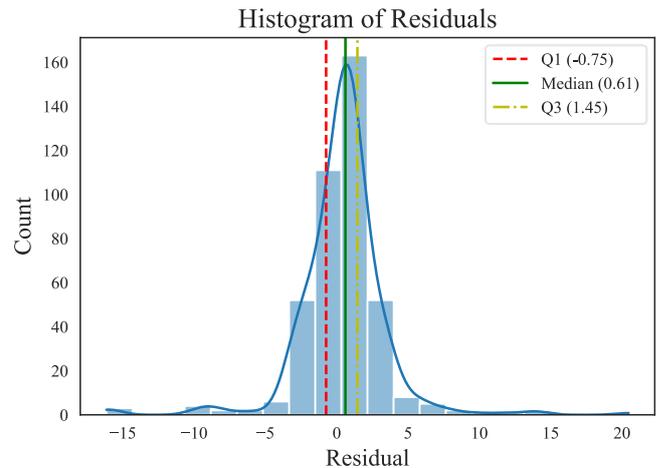

Fig. 7. Histogram of residuals for the proposed algorithm

The performance of the proposed algorithm's prediction is shown in Fig.8. The estimated friction force estimated by the algorithm and computed using the equation of motion in different operating conditions aligns on a line with a slope of 1 and an intercept of 0, showing the algorithm's precision in identifying the relationships in the data.

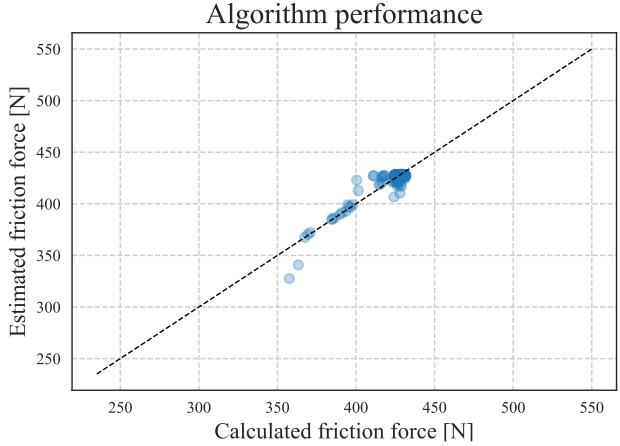

Fig. 8.   Estimated friction force vs. calculated friction force

The residuals histogram for the LuGre analytical model is displayed in Fig.9, and it can be seen that the residuals have a positive skewness (left-skewed) distribution. This indicates that many of the residuals have values below the distribution mean, while a small percentage of the residuals have values significantly above the mean. The tendency of the LuGre model to underestimate is indicated by the concentration of many residuals to the left of zero. This could indicate that the LuGre analytical model has poorly predicted actual values in some cases, leading to results lower than the friction derived from the equation of motion.

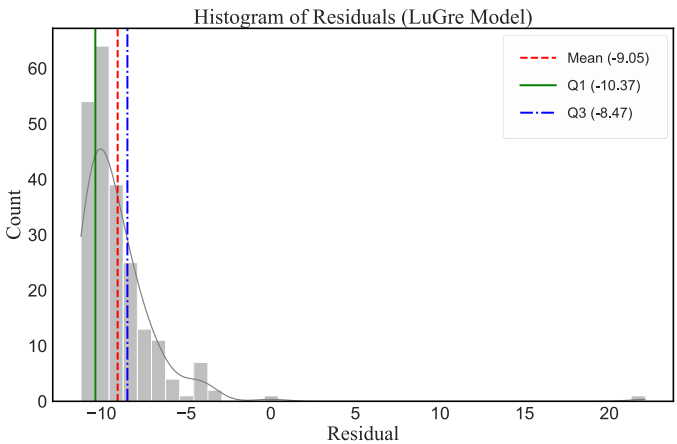

Fig. 9.   Histogram of residuals for the LuGre model

The computational cost of the proposed algorithm and model has been evaluated and is shown in the table III.

TABLE III.     COMPARISON OF COMPUTATIONAL COST BETWEEN THE LuGre MODEL AND THE PROPOSED ALGORITHM

| Test | Computational Cost |
|---|---|
| LuGre model | $8.71 \times 10^{-5}$ seconds |
| Algorithm | $1.51 \times 10^{-3}$ seconds |

## V.  CONCLUSION

This study introduced a hybrid data-driven algorithm that integrates Long Short-Term Memory (LSTM) network with Random Forest model to estimate nonlinear friction force in hydraulic cylinders online. The algorithm achieved high accuracy, with a mean absolute error (MAE) of 5.57% under variable load conditions, and outperformed the LuGre model, whose error reached 28.5% under similar circumstances. The exceptional performance of the proposed method stems from its ability to dynamically adapt to varying operating conditions and its robust architecture.

The LSTM network captured temporal dependencies and nonlinear patterns in time-series data by leveraging its memory cells, enabling precise modeling of friction dynamics over time. Additionally, the Random Forest model contributed to the robustness of the decision-making process, ensuring resistance to overfitting and noise. The algorithm's architecture was optimized using the Optuna framework, which systematically tuned key hyperparameters, such as the number of LSTM layers and nodes, dropout rates, and the number of Random Forest estimators, to minimize the MAE while maintaining computational efficiency.

Training dataset from a hydraulic test setup, incorporating pressure signals from piston chambers and velocity calculations under diverse conditions, provided a comprehensive basis for training and validation. Preprocessing steps, such as moving average filtering, reduce noise in the signals, further enhancing the model's reliability. Unlike the LuGre model, which struggles to adapt to external load changes without parameter re-identification, the hybrid algorithm demonstrated dynamic adaptability, effectively capturing the nonlinear friction characteristics across various operating conditions.

The proposed algorithm also proved computationally efficient, with an estimation time of 1.51 milliseconds, making it well-suited for real-time applications. This efficiency, combined with its accuracy and adaptability, highlights the algorithm's potential for deployment in industrial hydraulic systems, where precise friction force estimation is critical. Overall, integrating advanced machine learning techniques, optimized design, and systematic data preparation enabled the proposed algorithm to deliver superior performance, setting a new benchmark for real-time friction force estimation in hydraulic systems.

### AI USAGE STATEMENT

The authors declare that they used artificial intelligence tools only to improve and edit the text of the article.